\documentclass[11pt,letterpaper]{article}
\usepackage{cogsys}
\usepackage[T1]{fontenc}
\usepackage{times}
\usepackage[pdftex]{graphicx} 
\usepackage{amsmath,amssymb} 
\usepackage{array} 

\usepackage{natbib}
\cogsysheading{X}{20XX}{1-6}{X/20XX}{X/20XX}

\ShortHeadings{Diffusion Models and Concept Formation}
              {Z.\ Wang, K.\ Singaravadivelan, and C.\ J.\ MacLellan}

\begin{document}

\title{Diffusion Models and Concept Formation}

\author{Zekun Wang}{zekun@gatech.edu}
\author{Karthik Singaravadivelan}{ksingara3@gatech.edu}
\author{Christopher J.\ MacLellan}{cmaclell@gatech.edu}
\address{College of Computing, Georgia Institute of Technology,
         Atlanta, GA 30332 USA}
\vskip 0.2in

\begin{abstract}
Humans organize knowledge into a taxonomy of concepts with nested levels
of abstraction and a \emph{basic level} at which people recognize and name
objects with the least cognitive effort.
Cobweb is a classic cognitive account of this ability, an incremental
learner that builds a probabilistic concept hierarchy by maximizing
category utility. We argue that diffusion models, although
designed for image synthesis, implicitly perform the same computation.
The noisy marginals of a diffusion model are Gaussian smoothings of the
data distribution, and the modes of these marginals form a hierarchy that
corresponds to a Cobweb tree of probabilistic prototypes in four respects.
Both are hierarchical density models, both are hierarchical-Bayesian
models with Gaussian prototypes, both treat categorization as
score-following that reduces uncertainty, and in both a basic level
emerges. We locate this basic level for a diffusion model at an
intermediate noise level, where recent analyses show that the reverse
process commits to the class identity of a sample.
The two models differ mainly in how they represent and learn the taxonomy. Cobweb learns a discrete
tree incrementally, whereas a diffusion model encodes a continuous,
interpolable hierarchy in a single learned score field fit to the data distribution. We test the
correspondence on MNIST and Fashion-MNIST by recovering the diffusion
hierarchy through mode-finding and comparing the basic levels of the two
models. This reframes diffusion as a cognitive
model of concept formation and offers Cobweb a continuous, scalable
instantiation.
\end{abstract}

\section{Introduction}

Humans naturally organize their knowledge into nested levels of
abstraction. A collie is a kind of dog, a dog a kind of mammal, a mammal a
kind of animal. Within these hierarchies, one level supports recognition
with the least cognitive effort. This \emph{basic level} (e.g.,\ \emph{dog},
\emph{chair}) is the level people use by default, because it offers the
best trade-off between being informative about an instance's features and
being distinctive from other categories
\citep{rosch-1976-basicobjects,rosch-1975-familyresemblances}. Basic
categories maximize the joint predictability of cues and category
membership \citep{jones-1983-identifying}, and a computational account
of human categorization should explain why such a level emerges.

Cobweb \citep{fisher-1987-cobweb} offers a classic computational account.
It processes instances one at a time and incrementally builds a
probabilistic concept hierarchy, choosing at each step the restructuring
operation that most improves \emph{category utility}, an
information-theoretic measure of how much a partition increases the
predictability of instance features \citep{corter-1992-explaining}. The
resulting tree reproduces basic-level and typicality effects from the
psychology of categorization \citep{fisher-1988-basiclevel}. Recent work
has scaled the framework well beyond its original nominal-attribute
setting, to visual classification without catastrophic forgetting
\citep{barari-2024-cobweb4v}, to semantic retrieval
\citep{gupta-2025-retrieval}, to masked language modeling
\citep{lian-2025-masked}, and to deep taxonomic networks, which learn a
hierarchy of Gaussian prototypes in a latent space
\citep{wang-2025-taxonomic}.

In parallel, and largely independently, diffusion models
\citep{sohl-dickstein-2015-nonequilibrium,ho-2020-ddpm,song-2021-sde}
have become a leading approach to image generation. A diffusion
model learns to reverse a gradual noising process by estimating the
\emph{score} (the gradient of the log-density) of the data corrupted at
many noise levels. These models are typically studied as black-box
samplers and evaluated on sample quality, not as models of human concept
formation. Yet the object they learn, a family of progressively smoothed
data densities, has the structure of a concept hierarchy, a continuum of
representations of the data at different levels of abstraction.

This paper argues that the two lines of work describe members of the same
family of models. Our contributions are as follows.

\vskip 0.05in
\cbullet
We show that diffusion models and Cobweb belong to the same family of
hierarchical Bayesian density models with Gaussian prototypes, by making
four formal correspondences precise (Section~\ref{sec:core}).

\cbullet
We argue that diffusion models have a \emph{basic level}, an intermediate
noise level at which concepts are recognized with the least cognitive
effort.

\cbullet
We validate our claim by showing that hierarchies recovered through score-based mode-finding are consistent with Cobweb trees, and that their basic levels align with Cobweb's, with results on MNIST and Fashion-MNIST.
\vskip 0.05in

\noindent
The remainder of the paper reviews Cobweb, diffusion models, and
hierarchical generative clustering (Section~\ref{sec:background}), develops
their correspondences
(Section~\ref{sec:core}), presents the empirical validation
(Section~\ref{sec:experiments}), and discusses implications
(Section~\ref{sec:discussion}).

\section{Background}
\label{sec:background}

\subsection{Concept Formation with Cobweb}
\label{sec:cobweb}

Cobweb \citep{fisher-1987-cobweb} is an incremental and unsupervised model
of human concept formation. It processes a stream of instances and builds
a hierarchy of probabilistic concepts, where each concept node maintains a
summary of the instances it has adopted. As each instance arrives, Cobweb
sorts it down the tree from the root, and at every node it considers four
operations. It can add the instance to the best-fitting child, create a new
child for it, merge the two best children into one, or split the best child
and promote its children. It applies whichever operation most improves
category utility. In the original model each instance is a set of
nominal attribute--value pairs and a node stores a table of attribute
probabilities. Cobweb/3 extends the framework to continuous attributes by
storing, for each attribute, the mean and variance of a Gaussian
\citep{mckusick-1990-cobweb3}. We therefore take a node $c$ to carry a
prior $P(c)$ and a diagonal Gaussian over features,
$p(x \mid c) = \mathcal{N}(x; \mu_c, \sigma_c^2 I)$, treating features as
conditionally independent given the concept.

The measure that drives every decision is category utility. In its
information-theoretic form \citep{corter-1992-explaining}, adopted by
Cobweb/4V \citep{barari-2024-cobweb4v} and by deep taxonomic networks
\citep{wang-2025-taxonomic}, it scores a partition
$\{C_1,\dots,C_n\}$ of a node's instances by the expected reduction in the
uncertainty of feature values that knowing the child provides,

\begin{equation}
\mathrm{CU}(\{C_1,\dots,C_n\}) =
  \sum_{k=1}^{n} P(C_k)\,\bigl[\,H(A) - H(A\mid C_k)\,\bigr],
\label{eq:cu}
\end{equation}
where $H(A) = -\sum_{i}\sum_{j} P(A_i{=}V_{ij})\log P(A_i{=}V_{ij})$ is the
entropy over attribute values. Expressing the measure in terms of entropy
is what allows arbitrary distributions, such as Gaussians, to be plugged
in, since one only needs a closed-form entropy. Writing $X$ for the random
variable collecting the attributes and $C$ for the concept an instance is
assigned to, category utility is precisely the mutual information between
features and concept \citep{wang-2025-taxonomic},
\begin{equation}
\mathrm{CU} = I(X;C) = H(X) - \sum_{k=1}^{n} P(C_k)\,H(X\mid C_k).
\label{eq:cu-mi}
\end{equation}
A partition has high category utility when its clusters have low internal
entropy (predictable features) yet collectively account for the variation
in the parent (distinctiveness).

Although it was introduced decades ago
\citep{fisher-1987-cobweb,gennari-1989-incremental} and has long been
studied as a model of human categorization \citep{iba-2011-cobweb}, Cobweb
has seen renewed development, much of it aimed at scaling the framework past
nominal attributes and small symbolic datasets. Cobweb/4V learns concept
hierarchies over images and resists catastrophic forgetting in continual
learning, without replay buffers or regularization
\citep{barari-2024-cobweb4v}, and a follow-up traces this robustness to its
closed-form, information-theoretic updates rather than to gradient descent
\citep{barari-2025-robustness}. Concept-formation variants induce language
models and perform masked word prediction at scale
\citep{maclellan-2022-language,lian-2025-masked}, a Cobweb tree over
sentence embeddings supports coarse-to-fine semantic retrieval
\citep{gupta-2025-retrieval}, and, as a cognitive model, Cobweb accounts for
human-like incremental category learning \citep{lian-2024-cobweb}. Most
relevant here, recent work connects Cobweb with neural learning. Taxonomic
networks treat the concept hierarchy as a shared representation for pairing
symbolic and neural systems \citep{wang-2025-taxonomic-networks}, and deep
taxonomic networks recast it as a deep generative model, learning a
tree-structured mixture-of-Gaussians prior whose training objective
coincides with category utility \citep{wang-2025-taxonomic}. The latter
connects Cobweb directly to the generative latent-variable models we review
in Section~\ref{sec:hgc}.

\subsection{Diffusion Models}
\label{sec:diffusion}

A diffusion model is defined by a forward process that gradually corrupts
data with Gaussian noise and a learned reverse process that removes it
\citep{ho-2020-ddpm,song-2021-sde}. The forward process sets
$x_t = \sqrt{\bar\alpha_t}\,x_0 + \sqrt{1-\bar\alpha_t}\,\varepsilon$ with
$\varepsilon\sim\mathcal{N}(0,I)$ and a decreasing schedule
$\{\bar\alpha_t\}$, so that the distribution of $x_t$ is the data
distribution convolved with a Gaussian kernel, a kernel density estimate
of the data at bandwidth $\sigma_t := \sqrt{1-\bar\alpha_t}$,
\begin{equation}
p_t(x_t) = \int p_{\text{data}}(x_0)\,
  \mathcal{N}\!\bigl(x_t;\sqrt{\bar\alpha_t}\,x_0,(1-\bar\alpha_t)I\bigr)\,dx_0 .
\label{eq:marginal}
\end{equation}
The amount of corruption is summarized by the signal-to-noise ratio,
$\mathrm{SNR}(t) = \bar\alpha_t / (1-\bar\alpha_t)$, which decreases
monotonically from high values (almost-clean data, fine detail visible) at
small $t$ to near zero (almost pure noise) at large $t$
\citep{kingma-2021-vdm}.

Rather than learn $p_t$ directly, a diffusion model learns its
\emph{score}, the gradient of the log-density, which a standard result
identifies with a rescaled denoiser
\citep{hyvarinen-2005-scorematching,vincent-2011-dsm,song-2021-sde},
\begin{equation}
s_\theta(x_t,t) \approx \nabla_{x_t}\log p_t(x_t)
  = -\,\frac{\varepsilon_\theta(x_t,t)}{\sqrt{1-\bar\alpha_t}}.
\label{eq:score}
\end{equation}
The score also gives, through Tweedie's empirical-Bayes formula
\citep{robbins-1956-empiricalbayes,efron-2011-tweedie}, the posterior mean
of the clean datum given its corrupted version,
\begin{equation}
\hat x_0(x_t) = \mathbb{E}[x_0\mid x_t] = \frac{1}{\sqrt{\bar\alpha_t}}
  \bigl(x_t + (1-\bar\alpha_t)\,\nabla_{x_t}\log p_t(x_t)\bigr).
\label{eq:tweedie}
\end{equation}
A local mode of $p_t$ has vanishing score, so by Eq.~\ref{eq:tweedie} its
rescaled position lies on the clean data manifold,
$\hat x_0(x_t^{\star}) = x_t^{\star}/\sqrt{\bar\alpha_t}$
\citep{wang-2026-concept-discovery}. We recover such a mode from any datum
$x_0$ by noising it to level $t$ and then ascending the score until it
converges to a fixed point,
\begin{equation}
x_t^{(0)} = \sqrt{\bar\alpha_t}\,x_0 + \sqrt{1-\bar\alpha_t}\,\varepsilon,
\qquad
x_t^{(i+1)} = x_t^{(i)} + \eta\, s_\theta\bigl(x_t^{(i)},t\bigr).
\label{eq:modeascent}
\end{equation}
The noise level $t$ at which we do this fixes the level of abstraction of
the concept the datum is mapped to. A small $t$ returns a near-instance
prototype, and a large $t$ returns a broad class prototype. A diffusion
model thus provides a smoothed density at every noise level, together with
a score that locates the modes of that density and their local Gaussian
shape. These are the ingredients we need to connect it to Cobweb.

Throughout the paper we keep two indices distinct. The noise level is the
diffusion index $t$, whose smoothing bandwidth
$\sigma_t = \sqrt{1-\bar\alpha_t}$ grows with $t$, and the level of
abstraction is a position in the concept hierarchy that runs from fine to
coarse. A higher noise level gives a larger bandwidth, and a larger
bandwidth gives a coarser level of abstraction. We use \emph{noise level}
for the diffusion index and \emph{level of abstraction} for the hierarchy.

\subsection{Hierarchical Clustering}
\label{sec:hgc}

A parallel and largely separate literature builds hierarchies with
generative latent-variable models. The variational autoencoder learns a
latent representation of the data by maximizing a variational lower bound on
the likelihood \citep{kingma-2014-vae}. Placing a clustering prior on that
latent space turns it into a generative clustering model. VaDE uses a
Gaussian-mixture prior to recover flat clusters \citep{jiang-2017-vade}.
Making the prior hierarchical yields a tree of clusters rather than a flat
partition. TreeVAE grows a latent tree whose leaves are clusters
\citep{manduchi-2023-treevae}, DeepECT embeds a cluster tree inside an
autoencoder \citep{mautz-2020-deepect}, and, in the Bayesian-nonparametric
tradition, the nested Chinese restaurant process places a prior over trees
of unbounded depth \citep{blei-2010-ncrp}. Most relevant here, the deep
taxonomic network learns a binary-tree mixture-of-Gaussians prior and shows
that maximizing its evidence lower bound is equivalent to maximizing
Cobweb's category utility \citep{wang-2025-taxonomic}, placing Cobweb and
deep generative clustering on a common footing.

These models inherit a long tradition. Classical hierarchical clustering
builds a tree agglomeratively, merging the pair of clusters that least
increases within-cluster variance \citep{ward-1963-hierarchical}, or
divisively, by splitting from the top down. Density-based methods instead
treat clusters as the modes of a density and assign points to their basins
of attraction, as in mean shift \citep{comaniciu-2002-meanshift} and density
cluster trees. Cobweb is the cognitive-science member of this family, an
incremental, divisive conceptual clustering that summarizes each node with a
probabilistic prototype and chooses operations by category utility
\citep{fisher-1987-cobweb}. The deep latent-variable models extend this
tradition by learning the representation and the hierarchy jointly, rather
than clustering over fixed features.

We will argue that a diffusion model belongs to the same family, although
it represents the hierarchy implicitly rather than storing a tree. Its noisy
marginals are smoothed densities whose modes are cluster centroids at a
continuum of noise levels, recovered by mode-seeking on the learned score, which
is exactly the density-based view above. Two recent threads support this
reading. Work on concept discovery uses the modes of a diffusion model's
marginals directly as prototypes and composes them to generate new concepts
\citep{wang-2026-concept-discovery,du-2020-compositional,liu-2022-composable}.
And analyses of the diffusion process show that the structure it recovers
across noise levels is genuinely hierarchical, with a phase transition at an
intermediate noise level where coarse class identity is fixed while fine detail is
not \citep{sclocchi-2025-phasetransition,biroli-2024-dynamicalregimes,
ambrogioni-2025-thermodynamics}. These threads have developed separately
from Cobweb. Our contribution is to draw the correspondence between them
explicitly and to locate a basic level in the diffusion hierarchy.

\section{Diffusion Models as Hierarchical Concept Formation}
\label{sec:core}

We fix a common notation before drawing the correspondences. Throughout,
$x \in \mathbb{R}^d$ is a data point and $X$ the random variable it
realizes, $p_{\text{data}}(x_0)$ is the data distribution, $c$ is a concept
(a Cobweb node or a diffusion mode), and $C$ is the concept an instance is
assigned to. On the Cobweb side a
concept is a Gaussian $\mathcal{N}(\mu_c,\sigma_c^2 I)$ at a node of a
discrete tree. On the diffusion side a concept is a local mode
$x_t^{\star}$ of a smoothed marginal $p_t$, equipped with a local Gaussian
$\mathcal{N}(m_c,\Sigma_c)$ read off the score, as the next subsections
develop. The four correspondences are summarized in
Table~\ref{tab:correspondence} and argued in turn below. A final
subsection discusses where the two models differ.

\begin{table}[t]
\caption{The correspondence between Cobweb concept hierarchies and the
         implicit hierarchy of a diffusion model. Each row is argued in
         the indicated subsection of Section~\ref{sec:core}.}
\label{tab:correspondence}
\begin{center}
\begin{small}
\setlength{\tabcolsep}{4pt}
\renewcommand{\arraystretch}{1.15}
\begin{tabular}{@{}m{2.5in} c m{2.5in}@{}}
\hline
\abovespace\belowspace
\textbf{Cobweb} & & \textbf{Diffusion model} \\
\hline
\abovespace
tree depth (level of abstraction)
  & $\Longleftrightarrow$ &
noise level $\sigma_t$, equivalently $\mathrm{SNR}(t)$ (\S\ref{sec:density}) \\
node Gaussian $\mathcal{N}(\mu_c,\sigma_c^2 I)$
  & $\Longleftrightarrow$ &
mode Gaussian $\mathcal{N}(m_c,\Sigma_c)$ (\S\ref{sec:bayes}) \\
root-to-leaf categorization
  & $\Longleftrightarrow$ &
coarse-to-fine denoising (\S\ref{sec:score}) \\
\belowspace
basic level $\arg\max_{c\in\pi(x)} \mathcal{D}(c)$
  & $\Longleftrightarrow$ &
maximum-likelihood mode at noise $t^{\star}$ (\S\ref{sec:basiclevel}) \\
\hline
\end{tabular}
\end{small}
\end{center}
\end{table}

\subsection{Both Are Hierarchical Density Models}
\label{sec:density}

A Cobweb tree is a hierarchy of densities. Each node summarizes its
subtree with a probabilistic prototype, leaves capture individual
instances, and ancestors describe progressively broader regions of feature
space \citep{fisher-1987-cobweb}. Because Cobweb sorts
instances from the top down using its insert, create, merge, and split
operators, it can be read as a divisive procedure that refines coarse regions of feature space into finer ones, keeping every node a good summary of the instances below it.

A diffusion model produces the same kind of hierarchy, indexed by a
continuous noise level rather than by tree depth. For a finite dataset
$\{x_0^{(i)}\}_{i=1}^N$, Eq.~\ref{eq:marginal} is a Gaussian mixture with
one component per training point,
\begin{equation}
p_t(x_t) = \frac{1}{N}\sum_{i=1}^{N}
  \mathcal{N}\!\bigl(x_t;\sqrt{\bar\alpha_t}\,x_0^{(i)},(1-\bar\alpha_t)I\bigr),
\label{eq:gmm}
\end{equation}
a kernel density estimate whose bandwidth $\sigma_t$ grows with $t$. The
number of modes of this estimate falls monotonically as $\sigma_t$
increases. Fine, instance-level modes at small $t$ give way to broad,
class-level modes at large $t$. These progressively smoothed densities,
one at each noise level, form a family whose modes merge into a tree. A coarse mode at level $t'$ is the parent of
the finer modes at $t<t'$ that flow into it under score ascent, so depth
in the tree corresponds to the bandwidth $\sigma_t$. Finding the modes at a
fixed noise level is the mode-seeking problem that mean shift solves
\citep{comaniciu-2002-meanshift}, here driven by the learned score
(Eq.~\ref{eq:score}) rather than a fixed kernel, and following modes across
noise levels yields a cluster hierarchy in the spirit of agglomerative clustering
\citep{ward-1963-hierarchical} but read off a single learned density.
Empirically, modes of $p_t$ consolidate from instance-level to
object-level prototypes as $t$ grows \citep{wang-2026-concept-discovery},
and the structure that diffusion recovers across noise levels is hierarchical
\citep{sclocchi-2025-phasetransition}, closely
matching the tree Cobweb learns on the same data (see
Figure~\ref{fig:hierarchies}).

\subsection{Both Are Hierarchical Bayes with Gaussian Prototypes}
\label{sec:bayes}

The prototypes in both hierarchies are Gaussian, and in both the broader
concepts are mixtures of the narrower ones. A Cobweb node stores a diagonal
Gaussian $\mathcal{N}(\mu_c,\sigma_c^2 I)$ that pools the statistics of its
children. With $P(k\mid c)$ the weight of child $k$ under parent $c$, the
parent mean is the convex combination of the children's means,
\begin{equation}
\mu_c = \sum_{k} P(k\mid c)\,\mu_k, \qquad \sum_k P(k\mid c) = 1,
\label{eq:parent}
\end{equation}
and the law of total variance gives
$\sigma_c^2 = \sum_k P(k\mid c)\,[\,\sigma_k^2 + (\mu_k-\mu_c)^2\,]$, so a
parent is a broader Gaussian that subsumes its children and the hierarchy
is a tree-structured mixture of Gaussians \citep{mckusick-1990-cobweb3}.
Deep taxonomic networks impose a binary tree and learn the same pooling as
a single mixing weight, $\mu_c = \lambda\,\mu_{\text{left}} +
(1-\lambda)\,\mu_{\text{right}}$ with $\lambda\in[0,1]$, the two-child
special case of Eq.~\ref{eq:parent} \citep{wang-2025-taxonomic}. A diffusion model gives the same
object analytically. At a mode $x_t^{\star}$, Tweedie's formula
(Eq.~\ref{eq:tweedie}) places the prototype mean at
$m_c = x_t^{\star}/\sqrt{\bar\alpha_t}$, and the Jacobian of the denoiser
sets its covariance,
\begin{equation}
\Sigma_c = \frac{1-\bar\alpha_t}{\sqrt{\bar\alpha_t}}\,
  \nabla_{x_t}\hat x_0(x_t)\big|_{x_t = x_t^{\star}},
\label{eq:laplace}
\end{equation}
the posterior covariance of the clean datum, which at a mode is the Laplace
approximation of $p_t$ \citep{wang-2026-concept-discovery}. Each mode
thus carries a local Gaussian $\mathcal{N}(m_c,\Sigma_c)$, and the empirical
marginal of Eq.~\ref{eq:gmm} is a mixture of such Gaussians whose
components broaden and fuse as $t$ grows. The node Gaussians of Cobweb and
the mode Gaussians of a diffusion model therefore play the same role. Both
models explain a datum with a Gaussian component whose breadth encodes its
level of abstraction \citep{mckusick-1990-cobweb3,wang-2026-concept-discovery}.

\subsection{Both Maximize Information About the Data}
\label{sec:score}

The third correspondence is that both models pursue the same objective,
making a concept as informative as possible about the data it explains,
even though they pursue it by different mechanisms. In Cobweb the objective
is category utility, which by Eq.~\ref{eq:cu-mi} is the mutual information
$I(X;C)$ between the features and the concept. The feature entropy $H(X)$
is fixed, so maximizing $I(X;C)$ minimizes the residual uncertainty
$H(X\mid C)$, and a concept scores well when it leaves little to guess
about its members. A diffusion model optimizes the matching quantity for a
continuous hierarchy. Its training objective is a denoising error weighted
by the rate of change of the signal-to-noise ratio \citep{kingma-2021-vdm},
\begin{equation}
\mathcal{L}(\theta) = \tfrac{1}{2}\,\mathbb{E}_{x_0,\varepsilon}
  \int_0^1 w(t)\,\bigl\|x_0 - \hat x_\theta(x_t,t)\bigr\|^2\,dt,
\qquad w(t) = -\,\mathrm{SNR}'(t),
\label{eq:vdm}
\end{equation}
with $\hat x_\theta$ the denoiser of Eq.~\ref{eq:tweedie} and $w(t)\ge 0$
because the ratio decreases in $t$. By the I-MMSE relation
\citep{guo-2005-immse}, the best achievable value of this integrated error
is the mutual information the noisy observations carry about the clean
datum, and at the optimal denoiser it equals the datum's negative
log-likelihood exactly \citep{kong-2023-itd}. Minimizing Eq.~\ref{eq:vdm}
therefore drives the model to extract as much information about the clean
datum as the noise leaves available, at every noise level. Category utility
and the denoising objective are two forms of the same principle, each
making the concept retain as much information about the data as it can.

The two objectives are pursued by different search procedures, and this is
where the coarse-to-fine order appears. Cobweb sorts an instance from the
root down, at each node applying whichever of its four operators yields the
child partition of highest category utility, so it commits to broad
category structure first and refines toward the leaf. A diffusion model
reaches the same order by sampling, following the learned score from low
signal-to-noise (coarse, class-level structure) to high signal-to-noise
(fine, instance-level detail) \citep{song-2019-ncsn,karras-2022-edm}. The
score points in the direction the trained denoiser estimates, so each step
refines a coarse guess into a more specific one, the continuous analog of
Cobweb's root-to-leaf sort. Both models therefore share this objective and
this coarse-to-fine order, and differ in how they represent and search the
hierarchy, a difference we take up in Section~\ref{sec:differ}.

\subsection{A Basic Level of Greatest Distinctiveness}
\label{sec:basiclevel}

The basic level is the level of abstraction at which a learner obtains the
useful category most informative about its members, distinctive enough to
set them apart from the data as a whole, yet not so specific that it
collapses onto individual instances
\citep{rosch-1976-basicobjects,fisher-1988-basiclevel}. On the cognitive
science account the basic level is a frontier of concepts across the
hierarchy \citep{fisher-1988-basiclevel}. These concepts need not sit at a
uniform depth, and each is the concept of greatest distinctiveness along its
path, the one most informative about its own members.
We measure how well a concept meets this standard by its distinctiveness,
\begin{equation}
\mathcal{D}(c)
= \mathbb{E}_{x\sim p(x\mid c)}\bigl[\mathrm{pmi}(x;c)\bigr]
= D_{\mathrm{KL}}\!\bigl(p(x\mid c)\,\|\,p(x)\bigr) \ge 0,
\label{eq:kl}
\end{equation}
the expected pointwise mutual information of its members, which is the
Kullback--Leibler divergence of the concept from the data marginal and is
what information theory calls the \emph{specific surprise} of $c$
\citep{deweese-1999-specific}. Two related quantities place
$\mathcal{D}(c)$ in context (Table~\ref{tab:pmi}). Its pointwise ingredient,
the pointwise mutual information
$\mathrm{pmi}(x;c)=\log\bigl[p(x\mid c)/p(x)\bigr]$ of one datum and one
concept, is the evidence that $x$ belongs to $c$. Averaging $\mathcal{D}(c)$
in turn over concepts recovers the mutual information between data and
concepts,
\begin{equation}
I(X;C) = \sum_c p(c)\,\mathcal{D}(c) \ge 0,
\label{eq:cu-mi-local}
\end{equation}
the information-theoretic counterpart of category utility, with the sum
running over the concepts of one partition of the data. Distinctiveness
is the per-concept quantity between these two, and it is the one we use to
locate the basic level, by the same computation in Cobweb and in the
diffusion model.

\begin{table}[t]
\caption{Distinctiveness $\mathcal{D}(c)$, the quantity that locates the
         basic level, shown with its pointwise ingredient
         $\mathrm{pmi}(x;c)$ and its dataset-level aggregate $I(X;C)$.
         Distinctiveness is the expected pointwise mutual information of a
         concept's members. The pointwise term is signed, and
         $\mathcal{D}(c)$ and the mutual information are nonnegative.}
\label{tab:pmi}
\begin{center}
\begin{small}
\begin{tabular}{p{1.0in}p{1.95in}p{1.85in}}
\hline
\abovespace\belowspace
\textbf{Scope} & \textbf{Quantity} & \textbf{Reading} \\
\hline
\abovespace\belowspace
specific $x$ and $c$
  & $\mathrm{pmi}(x;c)=\log\dfrac{p(x\mid c)}{p(x)}$
  & evidence that $x$ belongs to $c$\\
\hline
\abovespace\belowspace
fixed $c$, over its members $x$
  & $\mathcal{D}(c)=\mathbb{E}_{x\sim p(x\mid c)}[\mathrm{pmi}]$
  & distinctiveness of a concept, the KL specific surprise \\
\hline
\abovespace\belowspace
over both $c$ and $x$
  & $I(X;C)=\sum_c p(c)\,\mathcal{D}(c)$
  & mutual information, the counterpart of category utility \\
\hline
\end{tabular}
\end{small}
\end{center}
\end{table}

Distinctiveness is defined at every concept, so a basic level exists in
any concept-formation system as the frontier where $\mathcal{D}$ peaks.
Along each root-to-leaf path $\pi(x)$, the concepts that contain
$x$ from the root down to its leaf, we take the concept of greatest
distinctiveness,
\begin{equation}
c_{\mathrm{basic}}(x) = \arg\max_{c\in\pi(x)} \mathcal{D}(c).
\label{eq:basic-frontier}
\end{equation}
This score is converging to zero near the root, where $p(x\mid c)=p(x)$. It need
not be monotone in depth, and in our experiments it reaches its maximum at
an intermediate level of abstraction rather than at the root or the leaves
(Figure~\ref{fig:basiclevel}). In Cobweb the concepts are explicit, and we read this level directly
off the learned tree (see
Figure~\ref{fig:basiclevelconcepts} for the basic-level prototypes of both
models).

In a diffusion model the concepts are implicit. There is no stored tree and
no closed form for $p(x\mid c)$, so the same level must be recovered
empirically, by passing held-out data through the trained model. Each datum
$x$ induces a categorization path $\pi(x)$, the sequence of local modes
from high noise (coarse) to low noise (fine), each carrying the local
Gaussian $\mathcal{N}(m_c,\Sigma_c)$ of Section~\ref{sec:bayes}. 
For a fixed datum, the marginal \(\log p(x)\) is constant across the concepts on its path. Thus, choosing the concept of greatest pointwise mutual information is the same as choosing the concept of greatest likelihood,
\begin{equation}
c^{\star}(x)
= \arg\max_{c\in\pi(x)} \mathrm{pmi}(x;c)
= \arg\max_{c\in\pi(x)} \bigl[\log p(x\mid c) - \log p(x)\bigr]
= \arg\max_{c\in\pi(x)} \log p(x\mid c).
\label{eq:maxll}
\end{equation}
With the local covariance taken to be diagonal,
\(\Sigma_c=\operatorname{diag}(\sigma_{c,1}^2,\dots,\sigma_{c,d}^2)\), the
likelihood decomposes into a sum over coordinates of a standardized data-fit
term and a log-variance term,
\begin{equation}
\log p(x\mid c)
= -\frac{1}{2}\sum_{j=1}^{d}\frac{(x_j-m_{c,j})^2}{\sigma_{c,j}^2}
\;-\; \frac{1}{2}\sum_{j=1}^{d}\log\!\bigl(2\pi\,\sigma_{c,j}^2\bigr).
\label{eq:gauss-ll}
\end{equation}
The first term favors concepts whose centroids lie close to \(x\) relative to their spread.
The second favors compact, low-variance concepts.  Along a hierarchical path, these competing pressures can make the likelihood peak at an intermediate level of abstraction.
We estimate $\mathcal{D}(c)$ in diffusion models from data by replacing the expectation in
Eq.~\ref{eq:kl} with an empirical mean over the members assigned to a
concept:
\begin{equation}
\mathcal{D}(c) \;\approx\;
  \frac{1}{N_c}\sum_{x\,\in\,\mathrm{members}(c)} \mathrm{pmi}(x;c),
\qquad N_c = \lvert\mathrm{members}(c)\rvert.
\label{eq:dhat}
\end{equation}

Thus, \(\mathcal{D}(c)\) can be measured from both models. 
In Cobweb we evaluate it at each node of the learned tree,
approximating the marginal $p(x)$ by a mixture over the nodes. In the
diffusion model we never form $p(x)$. The marginal cancels for a fixed
datum in Eq.~\ref{eq:maxll}, and once we average over the held-out set it
adds only a constant that is the same at every level, so the level at which
mean distinctiveness peaks is the level at which the mean held-out
log-likelihood peaks. 
This interior peak is a bias--variance tradeoff over levels of abstraction:
the finest concepts are nearly unbiased for the instances they store but
have high variance, predicting a held-out point poorly whenever it does not
fall near one of them, whereas coarser concepts are more biased, pulled
toward the mean of their members, yet vary less from one datum to the next.
The basic level is the abstraction that balances the two.
Read as a function of the level of abstraction, which is tree depth in
Cobweb and noise level in the diffusion model, $\mathcal{D}$ is zero at the
root and reaches its maximum at an intermediate level. The noise level $t^{\star}$ at
which the diffusion curve peaks is its basic level
(see Figure~\ref{fig:basiclevel}).

\subsection{Discrete versus Continuous Hierarchies}
\label{sec:differ}

The two models are not identical, however, and their differences are
instructive. Cobweb learns an explicit,
discrete tree, incrementally and one instance at a time, using symbolic
insert, create, merge, and split operators, and it stores one prototype
per node. A
diffusion model encodes a continuous hierarchy implicitly in a single
learned score field. There is no stored tree, only a family of smoothed
densities whose modes and their merging behavior define a hierarchy that
must be recovered by mode-finding. One consequence is that diffusion
prototypes are interpolable. A convex combination of two mode means,
$m(\lambda) = \lambda\,m_a + (1-\lambda)\,m_b$, traces a path of
intermediate prototypes, whereas Cobweb's nodes are
fixed and discrete. The trade-off is the familiar one between symbolic and
sub-symbolic representations. Cobweb offers interpretability, resistance to
catastrophic forgetting through sparse and local updates
\citep{barari-2025-robustness}, and genuine incremental learning. A
diffusion model offers continuity, scalability, and sample quality
\citep{karras-2022-edm}. Because the two describe the same
structure by different means, a natural direction is to bridge them, for
instance by pairing a neural density model with a symbolic taxonomy
\citep{wang-2025-taxonomic-networks,gupta-2025-retrieval}.

\section{Validating the Connection}
\label{sec:experiments}

The correspondence of Section~\ref{sec:core} makes empirical predictions.
We lay them out as two questions, each comparing a diffusion model's
implicit hierarchy, recovered by score-based mode-finding, against a Cobweb
hierarchy learned on the same data. We use image benchmarks shared by both
lineages. MNIST and Fashion-MNIST each contain 70{,}000 grayscale images of
size $28\times 28$ in ten classes, handwritten digits and clothing items
respectively, split into 60{,}000 training and 10{,}000 test images.
Both models are fit on the training images, with the Cobweb implementation following Cobweb/4V
\citep{barari-2024-cobweb4v}.
Cobweb's concepts are explicit, so we read its
hierarchy and basic level from the learned tree, whereas the diffusion
model's concepts are implicit and we recover its hierarchy and basic level
by passing the held-out test images through the trained model. The figures
below report results on MNIST and Fashion-MNIST.

\subsection{Does Mode-Finding Recover a Cobweb-Like Hierarchy?}

\begin{figure}[p]
\begin{center}
\setlength{\tabcolsep}{2pt}
\begin{tabular}{c@{\hskip 4pt}c}
\rotatebox{90}{\scriptsize\shortstack{Cobweb\\MNIST}} &
  \includegraphics[width=5.6in,height=1.50in]{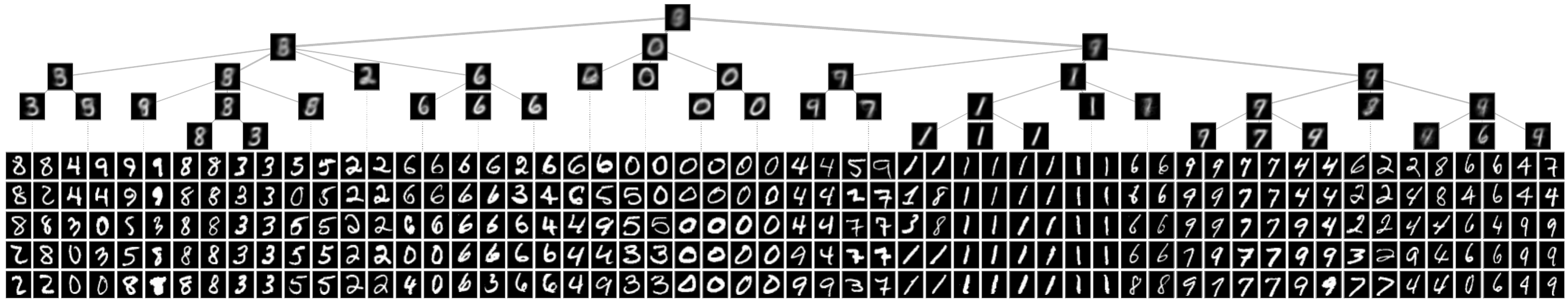} \\[6pt]
\rotatebox{90}{\scriptsize\shortstack{Diffusion\\MNIST}} &
  \includegraphics[width=5.6in,height=1.50in]{merge_path_taxonomy_mnist.pdf} \\[6pt]
\rotatebox{90}{\scriptsize\shortstack{Cobweb\\Fashion-MNIST}} &
  \includegraphics[width=5.6in,height=1.50in]{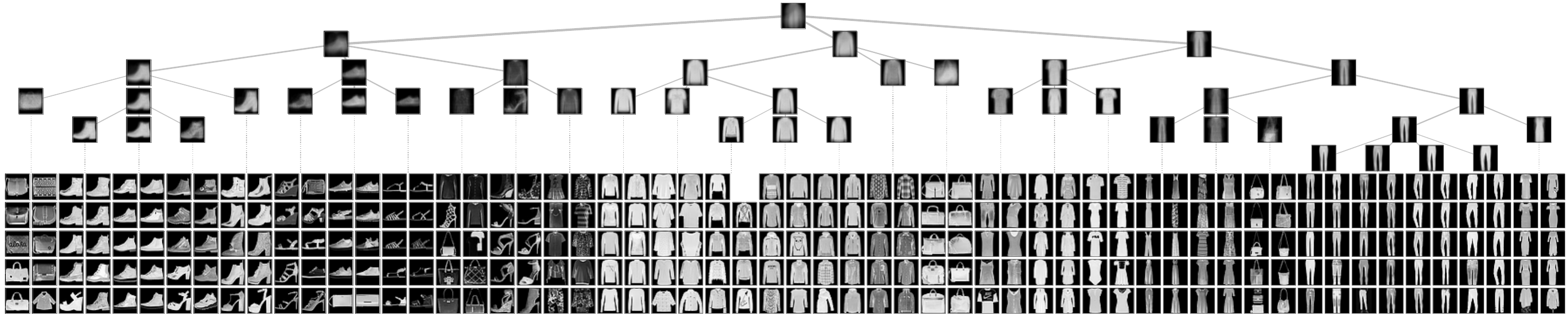} \\[6pt]
\rotatebox{90}{\scriptsize\shortstack{Diffusion\\Fashion-MNIST}} &
  \includegraphics[width=5.6in,height=1.50in]{merge_path_taxonomy_fashion_mnist.pdf} \\
\end{tabular}
\caption{Concept hierarchies recovered by Cobweb (rows~1 and~3) and by
  score-based mode-finding in a diffusion model (rows~2 and~4), on MNIST
  and Fashion-MNIST. The Cobweb rows show its concept taxonomy, with the
  root at the top, prototypes at internal nodes, and exemplars at the
  leaves. The diffusion rows show the taxonomy recovered by the merge-path
  procedure of Section~\ref{sec:experiments}. Modes found at each noise
  level are grouped into prototypes that merge into coarser parents as the
  noise level rises. Panel heights are normalized for display, and lower
  levels are shown only in part, since the branching factor grows toward the
  leaves. The two models recover structurally similar trees.}
\label{fig:hierarchies}
\end{center}
\end{figure}

The first prediction is that the hierarchy implicit in a diffusion model
is similar to the one Cobweb learns. We recover the diffusion hierarchy
bottom-up. Starting at a low noise level, we map every held-out test point
to a fine mode by noising it to that level and ascending the score
(Eq.~\ref{eq:modeascent}), and group modes that fall within a variance
threshold, taking each group's centroid to be the average of its modes.
Raising the noise level in steps, we agglomerate these groups into parents,
so that every parent encompasses the children beneath it and its centroid is
the average of theirs (Section~\ref{sec:bayes}). Stacking the levels gives a
taxonomy whose leaves are individual instances and whose root is a single
coarse mode, which we compare to a Cobweb tree on the same data
(Figure~\ref{fig:hierarchies}). 
Qualitatively, we see that both Cobweb and diffusion models create hierarchies in similar ways, merging similar stimuli (such as the 2s and 8s in the MNIST dataset, or the shoes and bags in the FashionMNIST dataset).

\subsection{Does the Diffusion Basic Level Match Cobweb's?}

The second prediction follows from Section~\ref{sec:basiclevel}. We score
every concept by its distinctiveness $\mathcal{D}$, estimated as the average
pointwise mutual information over the instances assigned to it
(Eq.~\ref{eq:dhat}), and read the basic level off the level at which
$\mathcal{D}$ peaks. In Cobweb the concepts are explicit, so we evaluate
$\mathcal{D}$ at each node of the learned tree, approximating the data
marginal by a mixture over the nodes. In the diffusion model the concepts
are implicit, so we recover them by routing the held-out test images through
the trained model along each datum's mode path (Eq.~\ref{eq:maxll}) and
aggregate, giving $\mathcal{D}$ as a function of the noise level.

\begin{figure}[t]
\begin{center}
\setlength{\tabcolsep}{6pt}
\begin{tabular}{cc}
\includegraphics[width=2.7in]{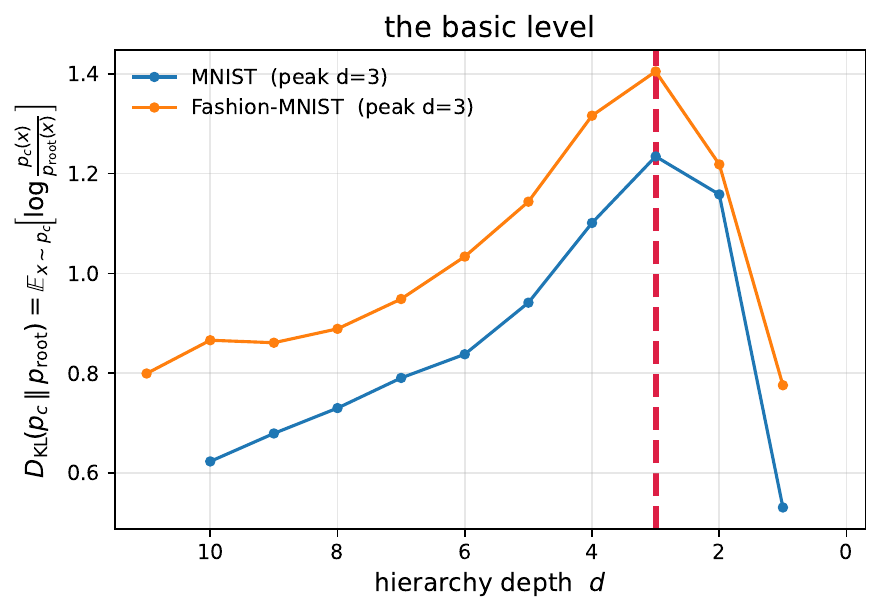} &
\includegraphics[width=2.7in]{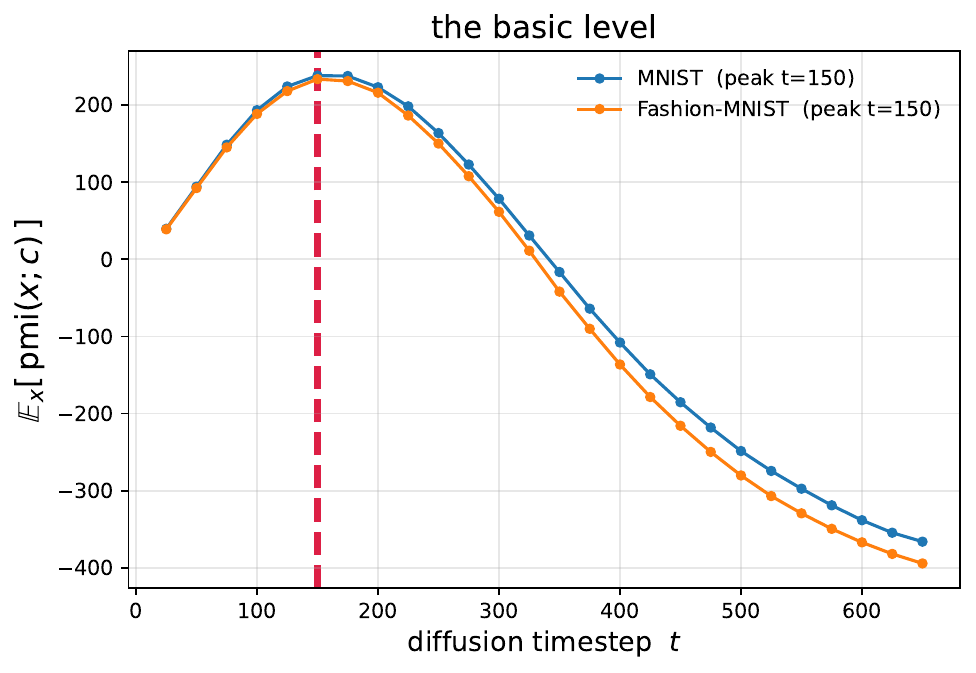} \\
{\small (a) Cobweb} & {\small (b) Diffusion} \\
\end{tabular}
\caption{Basic-level curves for the two models, each computed over both
  datasets. We plot the expected pointwise mutual information $\mathcal{D}$
  (Eq.~\ref{eq:kl}) against the level of abstraction, which is tree depth in
  (a) and noise level in (b). \textbf{(a)} In Cobweb,
  $\mathcal{D}$ against tree depth, peaking at depth three, with the root on the left. \textbf{(b)} In
  the diffusion model, $\mathcal{D}$ estimated along the mode paths against
  diffusion timestep, peaking near timestep~150. Each model shows a single
  interior maximum, the basic level (dashed line), rather than a maximum at
  the root or the leaves, on both MNIST and Fashion-MNIST.}
\label{fig:basiclevel}
\end{center}
\end{figure}

\begin{figure}[t]
\begin{center}
\includegraphics[width=5.4in]{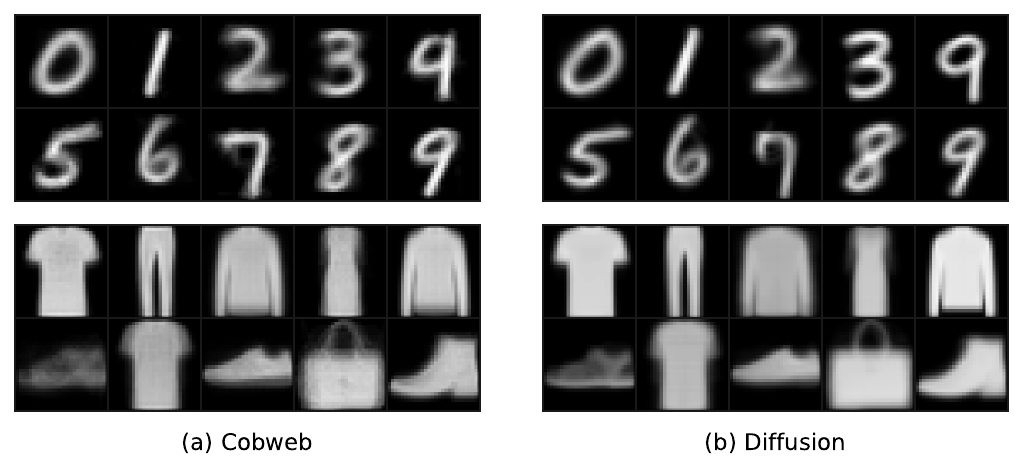}
\caption{The basic-level concepts at the peak levels of
  Figure~\ref{fig:basiclevel}. For each model we show ten prototypes from
  its basic level, one per class, with MNIST on the top rows and
  Fashion-MNIST on the bottom. \textbf{(a)} The Cobweb prototypes at depth
  three. \textbf{(b)} The diffusion prototypes recovered as the modes of the
  marginal at the peak timestep $t^{\star}\approx 150$ through Tweedie's
  formula. Each node is coarser than individual
  exemplars but finer than the whole dataset, and the matched pairs show the
  similarity of the basic level across the two models.}
\label{fig:basiclevelconcepts}
\end{center}
\end{figure}

Figure~\ref{fig:basiclevel} shows the resulting curves. Each has a single
interior maximum rather than a peak at the root or the leaves. The Cobweb
score peaks at depth three, and the diffusion score near timestep~150, for
both MNIST and Fashion-MNIST. The diffusion peak falls at an intermediate
noise level, consistent with the phase transition at which the reverse
process commits to a sample's class \citep{sclocchi-2025-phasetransition}.
Reading the concepts off these peaks makes the correspondence concrete.
Figure~\ref{fig:basiclevelconcepts} shows the basic-level prototypes of
each model. These are the Cobweb nodes at depth three and the diffusion modes
of the marginal at the peak timestep, recovered through Tweedie's formula.
Both are object-level categories, coarser than individual
exemplars yet finer than the dataset as a whole, and they align with the
categories people treat as basic.

\begin{figure}[t]
\begin{center}
\includegraphics[width=5.4in]{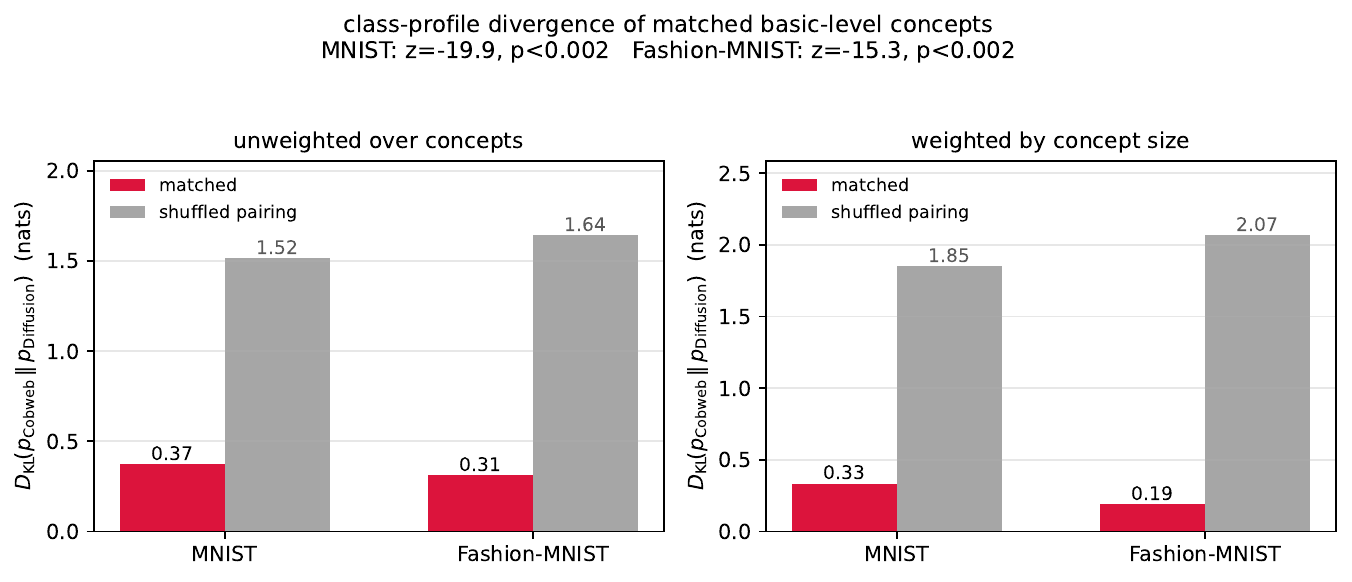}
\caption{Class-profile divergence of matched basic-level concepts.
  Kullback--Leibler divergence from each Cobweb basic concept's
  distribution over ground-truth classes to that of its greedily matched
  diffusion mode, \textbf{(a)} averaged over concepts and \textbf{(b)}
  weighted by concept size, against the same statistic when the pairing is
  shuffled. Matched concepts diverge three to seven times less than chance
  on both MNIST and Fashion-MNIST.}
\label{fig:matchingkl}
\end{center}
\end{figure}

In addition to showcasing this qualitative similarity, we conduct a
quantitative evaluation of the similarity of the basic level in both
Cobweb and diffusion models, showing results in Figure~\ref{fig:matchingkl}. Utilizing a greedy matching scheme of
basic-level concepts across the two models, we ask whether a Cobweb concept
and its diffusion counterpart are the same concept. The diffusion
frontier at $t^{\star}$ is the finer of the two (150 modes against 93
Cobweb concepts on MNIST, 108 against 64 on Fashion-MNIST), so we
repeatedly pair the closest unmatched Cobweb concept and diffusion mode by
prototype distance until every Cobweb concept has its best partner, then
fold each remaining mode into the group of its nearest partner and merge
the group into one concept. Each concept induces a distribution over
ground-truth classes, and Figure~\ref{fig:matchingkl} reports the
KL divergence from each Cobweb concept's distribution to
its matched mode's vs. when the pairing is
shuffled. Matched pairs diverge by $0.45$ nats on MNIST and $0.42$ on
Fashion-MNIST, and by $0.36$ and $0.28$ when concepts are weighted by
size, against $1.5$ to $2.1$ nats under shuffling.

Together, these results support our central claim that diffusion models and
Cobweb belong to the same family of models, recovering the same concept
hierarchy and the same basic level on MNIST and Fashion-MNIST.

\section{Discussion and Conclusion}
\label{sec:discussion}

We have argued that diffusion models and Cobweb belong to the same family
of models. Both are hierarchical density models, with tree depth in Cobweb
playing the role that noise level plays in diffusion. Their concepts are
Gaussian prototypes under a hierarchical Bayesian prior, stored at tree
nodes in one case and read off the modes of the smoothed marginals in the
other. Each categorizes by following a score toward the concept that most
reduces uncertainty about the instance. And in each, a basic level emerges
at which categorization takes the least cognitive effort, which we locate
for a diffusion model at the intermediate noise level where the reverse
process commits to a sample's class. Section~\ref{sec:experiments} tests
these claims. We recover the implicit hierarchy by mode-finding, then
compare its shape and its basic level against Cobweb's, and the two align
on MNIST and Fashion-MNIST. 

The correspondence has two implications for the cognitive science
community. First, diffusion models offer cognitive science a bridge to
deep learning. Cognitive models of
categorization have mostly been developed on small, hand-coded domains,
while modern generative models train on natural images at scale but are
treated as black boxes. The concept formation reading connects the two.
Prototypes, graded typicality, and the basic level all acquire
precise computational analogs inside a pretrained diffusion model, namely
the modes and
the noise level of maximal distinctiveness. Hypotheses about human categorization can therefore be
posed and tested on naturalistic stimuli without building a bespoke
cognitive architecture, and the results return to the machine learning
side as cognitive interpretations of what these models have learned. 

Second, this study is a proof of concept for studying symbol acquisition
in neural models. With diffusion models, mode-finding extracts discrete, stable prototypes
from a continuous density, and these proto-symbols compose, in that
combining their densities yields instances that satisfy several concepts
at once \citep{wang-2026-concept-discovery,du-2020-compositional,
liu-2022-composable}. Text-conditioned diffusion models already pair such
densities with words \citep{ho-2022-cfg}. A natural next study is to ask
how a word comes to pick out a mode of the density, at which level of the
hierarchy words attach most readily, and whether that level is the basic
level identified here. Broadly, symbolic and neural systems can be combined in many ways \citep{kautz2022third}, and the correspondence we draw shows one concrete route, pairing a symbolic taxonomy with a neural density model.

The limitation of our approach lies in how the hierarchy is indexed. We inherit the noise
schedule of diffusion practice, typically 1000 uniformly spaced
timesteps whose spacing was tuned for sample quality rather than for
cognitive plausibility \citep{ho-2020-ddpm,karras-2022-edm}. Equal steps
in timestep are not equal steps in abstraction. The mapping from timestep
to signal-to-noise ratio is nonlinear \citep{kingma-2021-vdm}, so the
recovered hierarchy resolves some ranges of abstraction finely and others
coarsely, and the basic level is reported in units that carry no cognitive
meaning of their own. Future work should explore cognitively inspired
schedules that place noise levels at human gradations of abstraction, for
example calibrating successive levels to subordinate, basic, and
superordinate structure \citep{rosch-1976-basicobjects}, or spacing them
by equal increments of category utility rather than of time. Indexing the
hierarchy by level of abstraction rather than by timestep would make the
correspondence with human concept hierarchies exact in form as well as in
structure.

\begin{acknowledgements}
\noindent
We would like to thank Pat Langley for discussions.

\end{acknowledgements}

\vspace{-0.25in}

{\parindent -10pt\leftskip 10pt\noindent
\bibliographystyle{cogsysapa}
\bibliography{references}

}

\end{document}